\documentclass[10pt,twocolumn,letterpaper]{article}

\usepackage{iccv}              % To produce the CAMERA-READY version
\usepackage[accsupp]{axessibility}
\usepackage{makecell}
\usepackage{algpseudocode}
\usepackage{algorithm}
\usepackage{caption}
\usepackage{subcaption}
\usepackage{array}
\usepackage[table]{xcolor}
\usepackage{colortbl}

\definecolor{iccvblue}{rgb}{0.21,0.49,0.74}
\usepackage[pagebackref,breaklinks,colorlinks,allcolors=iccvblue]{hyperref}

\def\paperID{*****} % *** Enter the Paper ID here
\def\confName{ICCV}
\def\confYear{2025}

\title{A Multi-purpose Tracking Framework for Salmon Welfare Monitoring in Challenging Environments}

\author{
Espen Uri Høgstedt$^{1}$ \quad
Christian Schellewald$^{2}$ \quad
Annette Stahl$^{1}$ \quad
Rudolf Mester$^{1}$ \\
$^{1}$Norwegian University of Science and Technology, Norway\\
$^{2}$SINTEF Ocean, Norway\\
{\tt\small espen.b.hogstedt@ntnu.no, christian.schellewald@sintef.no}, \\ 
{\tt\small annette.stahl@ntnu.no, rudolf.mester@ntnu.no}
}

\begin{document}
\maketitle
\begin{abstract}
Computer Vision (CV)-based continuous, automated and precise salmon welfare monitoring is a key step toward reduced salmon mortality and improved salmon welfare in industrial aquaculture net pens. 
Available CV methods for determining welfare indicators focus on single indicators and rely on object detectors and trackers from other application areas to aid their welfare indicator calculation algorithm. 
This comes with a high resource demand for real-world applications, since each indicator must be calculated separately. In addition, the methods are vulnerable to difficulties in underwater salmon scenes, such as object occlusion, similar object appearance, and similar object motion. 
To address these challenges, we propose a flexible tracking framework that uses a pose estimation network to extract bounding boxes around salmon and their corresponding body parts, and exploits information about the body parts, through specialized modules, to tackle challenges specific to underwater salmon scenes. Subsequently, the high-detail body part tracks are employed to calculate welfare indicators.
We construct two novel datasets assessing two salmon tracking challenges: salmon ID transfers in crowded scenes and salmon ID switches during turning. Our method outperforms the current state-of-the-art pedestrian tracker, BoostTrack, for both salmon tracking challenges.
Additionally, we create a dataset for calculating salmon tail beat wavelength, demonstrating that our body part tracking method is well-suited for automated welfare monitoring based on tail beat analysis. 
Datasets and code are available at https://github.com/espenbh/BoostCompTrack.
\end{abstract}    
\section{Introduction}
\label{sec:intro}
Enhancing the welfare of salmon in industrial net pens necessitates a detailed understanding of the health status of the fish. This information has traditionally been extracted by manual observation of sampled fish, which limits the quantity and quality of the data obtained. In recent years, it has been demonstrated that computer vision methods can be used to extract several welfare indicators \cite{Hogstedt2025, Xiao2015, Liu2014, Gupta2022}, offering new, more precise, and automated methods for salmon monitoring. Although promising, a number of challenges limit the real-world applicability of the proposed methods.
\begin{figure}
    \centering
    \includegraphics[width=\linewidth]{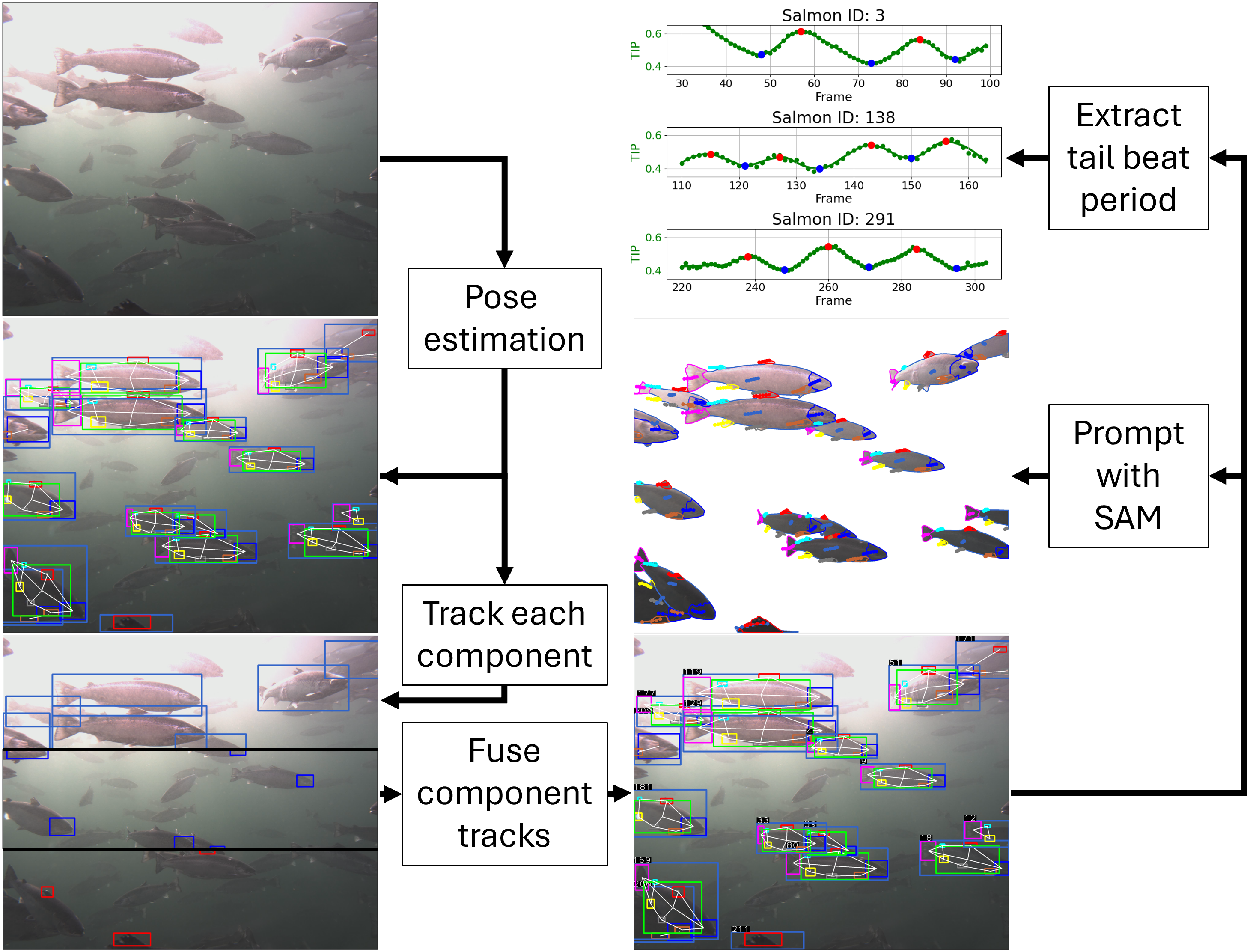}
    \caption{A visualization of our proposed pipeline \textit{BoostCompTrack}. Bounding boxes around salmon and salmon body parts are detected by a top-down pose model, before each salmon component is tracked separately and fused in a way that prevents certain challenges in underwater salmon tracking. The resulting tracks of salmon and salmon body parts can be leveraged for a diverse set of tasks, including segmentation mask generation with Segment Anything \cite{Kirillov2023} 
    %%% NEW 1.12 %%%
    (as demonstrated in the middle image on the right) and tail beat wavelength estimation (as visualized in the top image on the right).}
    %%%
    \label{fig:pipeline}
\end{figure}
% First shortcoming, bad data.
Firstly, the demonstrated algorithms are often developed with "ideal" data from controlled experimental setups. In commercial aquaculture facilities, the data is considerably more challenging. Complete and partial object occlusion is common and objects (fish in this case) move and appear very similar. Typical underwater scenes show significant illumination variability, motion blur, marine snow, and color-cast effects. Moreover, labeled data are a scarce resource, and the non-rigid camera mounting introduces motion, while the uniform background renders conventional camera motion compensation methods ineffective.

Secondly, each welfare indicator (or small group of indicators) is calculated by its own dedicated pipeline, often consisting of multiple deep neural networks. This results in excessive computational load, especially when considering multiple indicators concurrently to achieve a more comprehensive welfare estimate. 

Finally, extracting welfare indicators is usually heavily dependent on object localization and object tracking, yet available publications on salmon tracking and welfare monitoring tend to use object detectors and trackers from other application areas with little to no adaptation \cite{Nygaard2022, Zebele2022, Saad2024, Hogstedt2025}.
This leads to two main issues. Firstly, these trackers consider a point or a bounding box for each object, which is often not sufficiently informative to separate fish in dense crowds.
Secondly, any welfare indicator that is associated with specific locations on the salmon body requires an additional module to extract these locations. This module is challenging to construct, as it needs to both delineate a body part, and distinguish which body part belongs to the salmon of interest and which belongs to nearby fish occluding that particular salmon.

Several animal tracking methods operate with object keypoints \cite{Lauer2022, Pereira2022}, which are sufficiently informative for some welfare indicators \cite{Hogstedt2025}, while others benefit from bounding boxes \cite{Schellewald2024, Gupta2022} or segmentation masks \cite{Zhang2020, Clark2025}. Another challenge with pose estimation models is that they rely on distinguishable object keypoint locations for accurate pose inference. Furthermore, for tracking association, keypoints are often less robust than bounding boxes, which provide a more stable representation for matching objects across frames.
%Furthermore, pose estimation models require distinguishable object keypoint locations, and keypoints are less suited than bounding boxes in aiding tracking association algorithms.

% Make the argument for tracking
To address the challenges posed by tracking salmon in real-world underwater scenes, we propose a flexible object localization and tracking framework that explicitly copes with the difficulties specific to underwater salmon scenes and provides detailed object information that can be directly used to determine welfare indicators.
To achieve this, we use a top-down pose estimation network that first detects bounding boxes enclosing each clearly visible salmon and subsequently determines finer bounding boxes around specific body parts.
A state-of-the-art bounding box tracker is utilized to track detected salmon components separately, before specialized modules are employed to merge the associations at the two different scales (salmon and body parts) together to individual object associations.
This fusion step addresses task-specific challenges in salmon tracking, including often observed track ID transfers in crowded scenes and track ID switches when salmon are turning.
Following the tracking, the temporal evolution of the tail is determined to compute the salmon tail beat wavelength, demonstrating also the applicability of our tracker for salmon welfare monitoring. Additionally, the body parts can be prompted with foundational models (such as Segment Anything \cite{Kirillov2023}) to obtain tracks of segmented fish, which are displayed in \cref{fig:pipeline}.

To evaluate our tracking method, we assembled two challenging salmon tracking datasets covering two underwater application cases.
The first targets automatic data collection for training individual fish appearance models, requiring the tracker to maintain identity consistency and avoid switching between individuals (i.e., prevent fish ID transfer).
Furthermore, it benefits from having no appearance model, as such a model could introduce bias into the collected dataset by favoring certain fish tracks based on the feature embeddings of this particular model.
Learning individual appearance models is crucial for long-term salmon re-identification and, consequently, for monitoring individual salmon welfare.
%Calculating individual appearance models is an important task for long-term salmon re-identification, necessary for individual salmon welfare monitoring.
To assess this task, we compiled the CrowdedSalmon (CS) dataset, containing densely distributed salmon schooling in front of the camera (\cref{fig:dataset}). 

%The second application case addresses the problem of joining both visual salmon sides into one salmon individual, allowing enhanced surveillance capabilities for salmon monitoring. 
The second application case addresses the challenge of associating observations of a salmon’s left and right sides, captured at different times as the fish turns, into a single individual identity. This enables improved surveillance and individual monitoring.
With a single camera, this can only be achieved by maintaining a salmon track during turning, which is challenging since the turning fish changes appearance and bounding box shape.
To support this use case, we developed the TurningSalmon (TS) dataset, which features more dynamic scenes with multiple salmon turning simultaneously
(\cref{fig:dataset}). Additionally, we constructed the TailbeatWavelength (TBW) dataset to evaluate our tracker’s ability to support salmon motion monitoring through tail beat analysis (\cref{fig:dataset}).

Our results show that 
%%% NEW %%%
the proposed method, \textit{BoostCompTrack} (BCT), 
%%%
effectively reduce salmon ID transfers in crowded scenes and maintain consistent tracking through turning motions in dynamic environments. Furthermore, the detailed body part trajectories generated by our method can be directly utilized for autonomous estimation of salmon tail beat wavelength. 

\section{Related work}
%The most common approaches to address salmon tracking have been to separate object detection and track association into two distinct steps (tracking-by-detection [TBD]) \cite{Hogstedt2025, Nygaard2022, Zebele2022, Saad2024, Schellewald2024, Atlas2023}, where object detection has been performed by YOLO \cite{Redmon2016}, Keypoint RCNN \cite{He2017} or joint detection and embedding models \cite{Wang2020}, while the association has been performed by pedestrian tracking frameworks such as StrongSort \cite{Du2023} or DeepSort \cite{Wojke2017}.
The most common approach to salmon tracking has been to separate object detection and track association into two distinct steps (tracking-by-detection, TBD) \cite{Hogstedt2025, Nygaard2022, Zebele2022, Saad2024, Schellewald2024, Atlas2023}. Object detection is typically performed using models such as YOLO \cite{Redmon2016}, Keypoint RCNN \cite{He2017}, or joint detection-and-embedding models \cite{Wang2020}, while track association is often handled by pedestrian tracking frameworks like StrongSORT \cite{Du2023} or DeepSORT \cite{Wojke2017}.

General animal tracking is a more established research area, often using keypoint-based pose estimation to improve object association and the utility of resulting tracks. In contrast, salmon tracking presents particular challenges and remains underexplored, with fewer dedicated approaches
\cite{Lauer2022, Pereira2022}.

Since earlier salmon tracking methods were largely adapted from pedestrian tracking approaches, state-of-the-art pedestrian trackers serve as strong baseline models for salmon tracking.
The best-performing methods on the popular MOT20 benchmark \cite{Dendorfer2020} rely on a simple TBD scheme with different enhancements, including an improved intersection over union (IoU) cost matrix \cite{Stanojevic2024}, a Siamese network-based Similarity Learning Module \cite{Wang2024}, pseudo-depth \cite{Liu2025} and adaptive hyperparameter tuning \cite{Shim2024}.

% Part-based tracking
Work on part-based tracking (tracking bounding boxes around parts of hierarchical objects) has been scarce, particularly after the boom of deep learning-based object detectors. Earlier works include using part-based tracking to explicitly handle partial occlusion \cite{Shu2012}, and tracking humans and body parts simultaneously, modeling each body part as a Gaussian, and learning body part relationships online \cite{Izadinia2012}.

\subsection{Tail beat frequency}
Tail beat frequency (or the inverse, tail beat wavelength) is an important biometric for assessing salmon welfare, as it relates to stress level \cite{Tong2024}, swimming mechanics, and fish malformation. Earlier methods for automated tail beat frequency estimation are usually based on birds-eye view cameras, where the length, tail angle, or tail position (relative to the fish body) of isolated fish individuals is employed as an indicator of tail beat state \cite{Sasaki2024, Clark2025, Xiao2015, Terayama2017}. To the authors' knowledge, no computer vision method for salmon tail beat monitoring in industrial-sized net pens using a lateral-facing monocular camera has been presented. 
%%% New 2.12 %%%
Given that lateral-facing cameras have already proven effective for various aquaculture tasks, such as salmon re-identification, respiration monitoring and wound detection \cite{Hogstedt2025, Schellewald2024, Gupta2022}, extending their use to tail beat analysis presents an opportunity to develop integrated, multi-purpose monitoring systems.
%%%%
\section{Proposed tracking and analysis approach}
\label{sec:method}

\subsection{Details on the assembled datasets}
\label{sec:dataset}
\begin{figure}
    \centering
    \includegraphics[width=\linewidth]{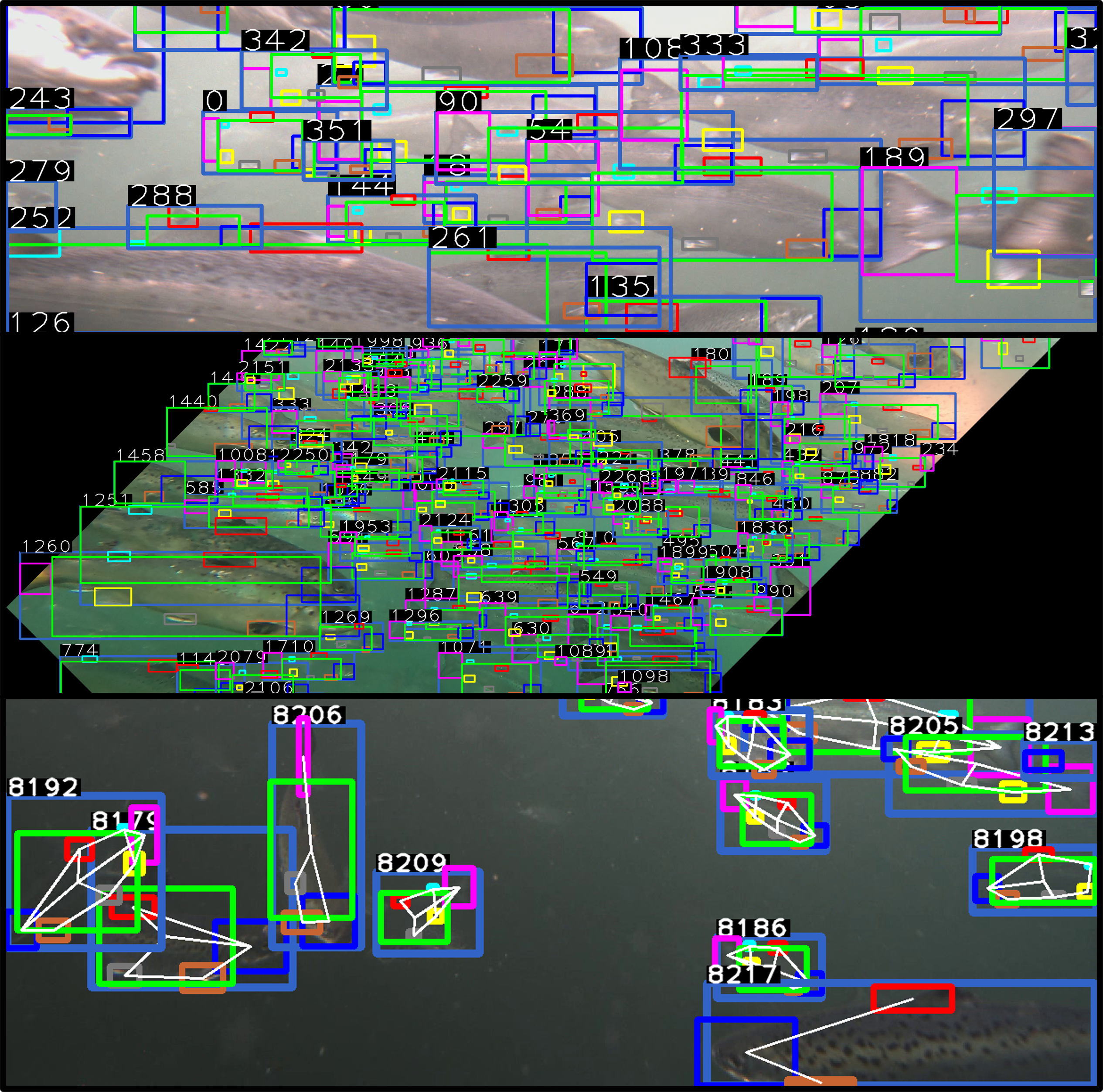}
    \caption{Annotated images from the TailbeatWavelength training dataset (top), CrowdedSalmon validation dataset (middle) and TurnSalmon validation dataset (bottom). 
    %%% NEW 2.17, 3.5 %%%
    All images from the CS dataset were rotated 135 degrees prior to processing to correct for the orientation of the raw video, which was captured at a 135-degree angle relative to the water surface.}
    %%%
    \label{fig:dataset}
\end{figure}

%%% NEW paragraph. 2.1, 2.3, 2.5, 2.6, 2.7, 2.8, 2.15 %%%
To robustly evaluate our framework, we constructed three datasets, CrowdedSalmon (CS), TurningSalmon (TS), and TailbeatWavelength (TBW), spanning different years, fish populations, and camera setups. See \cref{tab:datasets} for dataset specifications and \cref{fig:dataset} for visualizations. Each dataset split (row in \cref{tab:datasets}) is sampled from distinct video sequences, ensuring no data overlap between the splits. All datasets are derived from a single sequence, except TBW train, which includes frames from three video clips. Sampling intervals for the training datasets were 100 (CS), 1000 (TS), and 50 (TBW) frames, balancing data quantity and data diversity. The CS validation set was included in the training data of detectors used in generating the TS and TBW validation sets. This does not compromise evaluation integrity, as the TS and TBW validation sets remain entirely independent of the CS validation set. 
%%%

All training datasets were annotated with bounding boxes around all visible salmon and their non-occluded body parts (fins, head, and body, 
%%% NEW 2.16 %%%
see \cref{fig:TIP_example}).
%%%
A salmon was considered visible if we could reasonably locate most of the body parts of the fish after contrast and brightness adjustment. Each salmon component (salmon and salmon body parts) was assigned a frame-wise unique ID that linked salmon components together to form a full salmon.

The CS validation set was labeled similarly to the training sets, except that the specified IDs were unique globally, such that salmon number $i$ in frame $f$ was the same salmon as salmon number $i$ in frame $f+10$ (dataset images were sampled 10 frames apart
%%% NEW 2.4 %%%
to facilitate consistent identity assignment across frames, while ensuring adequate video coverage.)
%%%%%

%%% NEW paragraph. 2.7, 2.19 %%%
For the TS validation set, bounding boxes were automatically generated by a detection network (see \cref{sec:method_salmon_detection}) trained on the CS train, CS validation, and TS train datasets. For sufficiently long fish tracks, an early and a late frame in which the fish was clearly visible and the salmon bounding box well-positioned were manually annotated. These endpoints enable quantitative evaluation of tracking performance by measuring how often a tracker correctly links both ground truth detections within a track. Each track was also assigned one of five labels: \textit{Turning}, \textit{Straight}, \textit{Occluded}, \textit{Turning and Occluded}, or \textit{Background Occluded}.
%%%

%%% NEW shortened %%%
The TBW validation dataset was constructed by tracking all salmon components across a 1000-frame video using the BCT method in conjunction with object detector M5.  To increase the rate of complete salmon detections, the bpiou module (\cref{sec:method_salmon_detection}) was omitted. High-quality tracks were identified by excluding salmon instances with bounding box diagonals smaller than 200 pixels or occluded body parts, as well as object tracks with fewer than 50 consecutive frames. For each individual, only the longest continuous track was retained. Bounding boxes around salmon were annotated for each frame based on these selected tracks. Finally, extreme tail poses were labeled for salmon exhibiting at least four clearly discernible extremes. Individuals displaying turning or maneuvering behaviors were excluded, focusing the dataset on forward swimming patterns.
%%%

%%% NEW 2.5, 2.10, 2.13, 2.14 %%%
\begin{table}
\centering
\small
\setlength{\tabcolsep}{4pt}
\begin{tabular}{ |c|c|c|c|c|c|c|c|} 
\hline
Name & Imgs & Salmon & Boxes & Cam & Date & \makecell{net- \\ pen} \\ 
\hline
CS train & 31 & 1010 & 7627 & GoPro & 240418 & 1\\
CS val & 6 & 871 & 5814 & GoPro & 240418 & 1 \\ 
TS train & 8 & 679 & 5040 & Prop. 1 & 240502 & 2\\ 
TBW train & 46 & 1529 & 10615 & Prop. 2 & \makecell{230309, \\ 230311} & 1\\
\hline
TS val & 100 & 146 & N/A & Prop. 1 & 240508 & 2\\ 
TBW val & 1000 & 18 & N/A & Prop. 2 & 230309 & 1\\
\hline
\end{tabular}
\caption{Summary of CrowdedSalmon (CS), TurningSalmon (TS) and TailbeatWavelength (TBW) train and validation datasets. The GoPro camera recorded with a resolution of $2160 \times 3840$ pixels and with a frame rate of 30 FPS, while the raw image capable underwater proprietary cameras recorded at $1224 \times 1024$ (prop. 1) and $2448 \times 2048$ (prop. 2) pixels at 25 FPS. Each dataset (table row) was constructed from distinct video clips.
The datasets above the line are manually annotated with bounding boxes, while the datasets below the line are manually annotated with events and automatically annotated with bounding boxes. 
}\label{tab:datasets}
\end{table}
%%%

\subsection{Detectors}\label{sec:method_salmon_detection}
We employed the CS training set to train two types of salmon component detection networks: one for detecting bounding boxes and another for detecting keyboxes 
%%% NEW 2.18 %%%
(key area on the fish located by a bounding box). 
%%%
Both networks predict bounding boxes around all salmon components, but the keybox models additionally link each salmon’s body part boxes to its full-salmon bounding box. This allows for persistent tracking of individual components, even when some body parts are temporarily occluded.

%%% NEW shorten %%%
We trained five YOLO-based models \cite{yolov8_ultralytics}: four (M1-M4) on the CS train dataset, and one (M5) on all datasets with annotated bounding boxes (CS train, CS val, TS train and TBW train):
\begin{itemize}
    \item \textbf{M1}: yolov8n-pose (keybox) model, image size 640.
    \item \textbf{M2}: yolov8 detection model, image size 640.
    \item \textbf{M3}: yolov8m-pose (keybox) model, image size 1024.
    \item \textbf{M4}: yolov8m detection model, image size 1024.
    \item \textbf{M5}: yolov8m-pose (keybox) model, image size 1024.
\end{itemize}
%%%

%%% NEW 3.2 %%%
Given the limited dataset, we prioritized training coverage. Detector validation for M1-M4 was performed using a single, salmon-dense image from the CS training set, while M5 was validated with one image from each camera setup (GoPro, Prop. 1, and Prop. 2). The validation sets listed in \cref{tab:datasets} were withheld at this stage to ensure unbiased final tracker evaluation.
%%%

To construct the keybox detection networks, we formatted each body part bounding box as two keypoints (the anterodorsal corner and the posteroventral corner). All models were trained with the default Ultralytics settings, except for changing the random rotation parameter to 20 degrees, the scale parameter to 0.8 and the perspective parameter to 0.0001. The models were trained for 10000 epochs 
%%% NEW 3.3 %%%
(empirical observations indicated that optimal pose performance necessitated extensive training),
%%%
with a batch size of 8.

During inference, the confidence of each 
%%% NEW %%%
keybox
%%%
was calculated as the average between the occlusion values (keypoint confidence predicted by the pose models) of the two bounding box keypoints, multiplied by the salmon object confidence. The detector hyperparameters (confidence, keybox confidence, and salmon IoU threshold) used were those that maximized the F1 score between the detector results and the ground truth results on the CS train dataset. 

%In addition to the four models trained on the CS train dataset, we trained one keybox detection model (M5) on all datasets with annotated bounding boxes (CS train, CS val, TS train and TBW train). 
%%% NEW 3.2 %%%
%For this model, we used three images for detector validation, one from each camera setup (GoPro, prop. 1 and prop. 2).
%%%

We compare BCT with the original pedestrian tracking approach BoostTrack (BT) \cite{Stanojevic2024}, which achieves state-of-the-art performance on the MOT20 benchmark \cite{Dendorfer2020}. In our adaptation of BT for salmon tracking, we instantiate a separate BT tracker for each type of salmon component, ensuring that components of different types are not associated with one another.
Additionally, we attempted to benchmark our method against the widely used animal tracker DeepLabCut \cite{Lauer2022} by defining object keypoints as the centroids of the salmon body part bounding boxes. We were unable to obtain reliable and consistent detection results from the DeepLabCut framework after 1000 epochs of network training, so we excluded this method from further consideration.

The BCT method considers the complete salmon as one tracking unit with nine subtrackers, each identical to the object trackers of BT. The subtrackers of BCT are initially associated with the salmon component detections in the same way as BT. This yields a set of suggested tracker associations (one for each salmon component) for each detection. Initially, detections and trackers are associated based on the calculated salmon bounding box match (a salmon bounding box detection associated with a salmon bounding box tracker).

Subsequently, the \textit{TurnModule} and \textit{CrowdedModule} perform checks and subsequent improvements specifically targeting salmon ID transfers in crowded scenes and salmon ID switches for turning salmon.

The \textit{TurnModule} determines whether a salmon is turning by evaluating a counter $c$, which indicates turning when $c>0$. If a turning tracker lacks an associated detection, and its best-matching detection is not already assigned to another tracker, the match is accepted if the bounding box IoU overlap exceeds 0.05, even if below the original IoU threshold. The counter $c$ is modified by the following rules:
\begin{enumerate}
    \item \textbf{Increase}: The counter $c$ increases when the detection of the salmon suggests a frontal or posterior orientation, which is indicative of a turning maneuver. This increment is conditioned on the salmon not being adjacent to the image boundary and on the fulfillment of at least one of the following criteria: \begin{enumerate}
        \item The height of the salmon bounding box is larger than its width. 
        \item The distance between the head and tail fin is less than twice the distance between the dorsal and pelvic fins. 
    \end{enumerate}
    \item \textbf{Decrease}: The counter $c$ decreases when the detection of salmon indicates a lateral orientation, which is typically characterized by the visibility of most of the body parts. Specifically, $c$ is decreased if it was not increased in the current frame, and at least seven of the nine body parts are visible.
    \item \textbf{Range limitation}: By maintaining a value above zero despite some unmet increment conditions, the counter $c$ provides resilience against short-term turning detection inaccuracies. To prevent prolonged misclassifications of non-turning behavior, we bound $c$ between 0 and 10.
\end{enumerate}

If a salmon is not marked as turned, the \textit{CrowdedModule} checks whether the small body parts (the fins and the head) associations suggest that the initial salmon bounding box association needs correction. Specifically, it performs the following three tests and subsequent improvements:

\begin{enumerate}
    \item \textbf{Body part disagreement (bpdis)}: A salmon body part \textit{disagrees} with the salmon bounding box association if this body part is associated with a different fish than the salmon bounding box. If more small salmon body parts disagree than agree, and at least two components disagree, terminate the tracker associated with the salmon bounding box, and discard the corresponding detection.
    \item \textbf{No bodyparts (nobp)}: If no small salmon body parts are overlapping with the same detection as the salmon bounding box, terminate the tracker and discard its matched detection.
    \item \textbf{Low body part IoU (bpiou)}: If a small body part has a higher IoU overlap with a different detection than the one associated with the salmon bounding box, set the confidence of this body part detection to 0.
\end{enumerate}

The \textit{CrowdedModule} ensures early termination of problematic trackers, thereby preventing the propagation of poor-quality tracks. Once finalized, the resulting trajectories of individual salmon and their body parts can be analyzed further in subsequent applications, such as tail motion analysis, which we address in the following subsection.

\subsection{Tailbeat wavelength calculation}
To evaluate the tail beat wavelength, we analyzed observable/quantifiable oscillations of state values associated with the tail. We examined three ways of representing the salmon tail beat state.

\begin{figure}
    \centering
    \includegraphics[width=\linewidth]{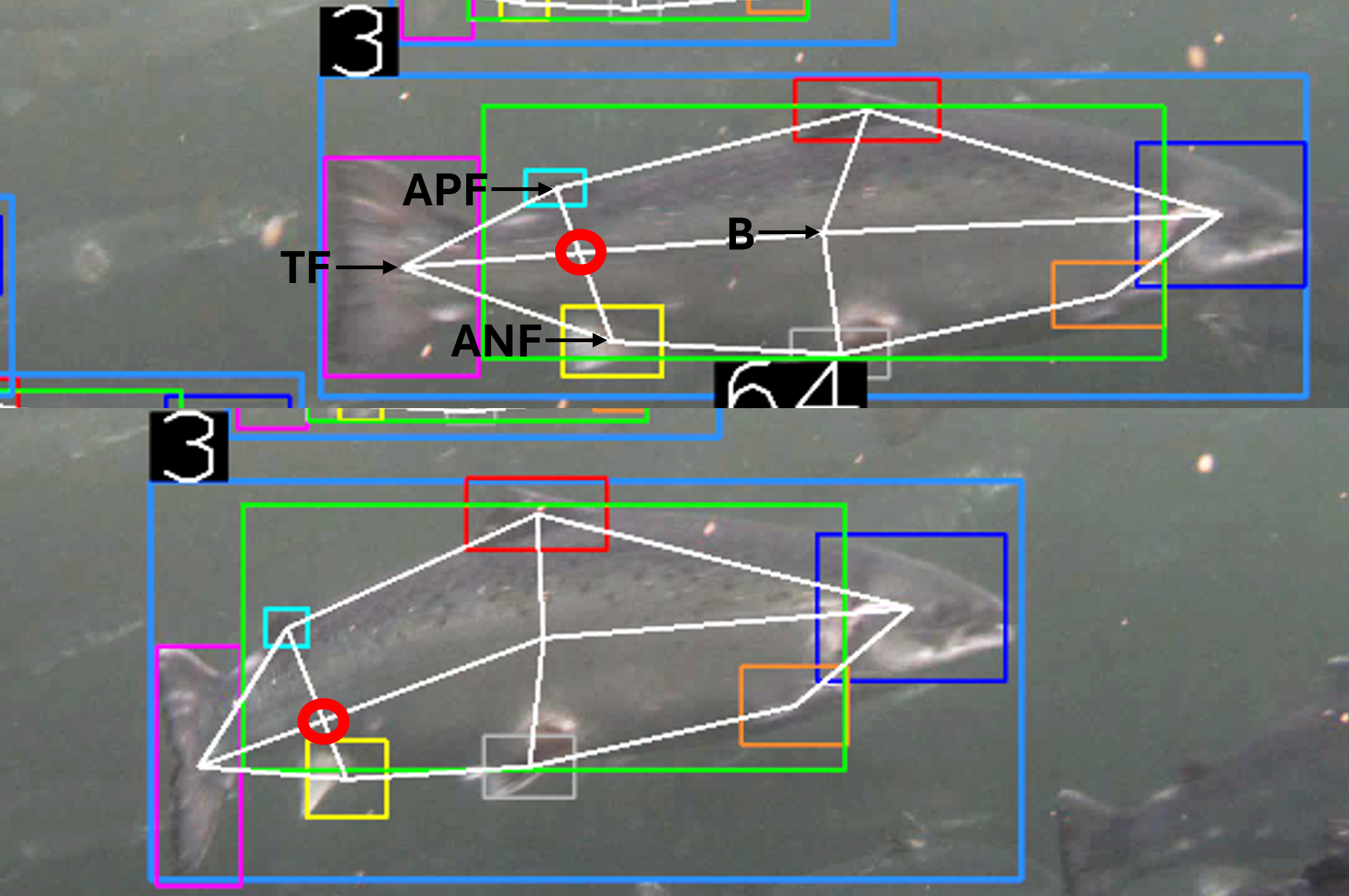}
    \caption{%%% NEW 1.10 %%%
    Two consecutive extreme salmon poses, annotated with bounding boxes around all salmon components, white lines between selected body parts
    %%%
    , and a red circle around the intersection point (IP). Additionally, the top frame is annotated with the points used to calculate IP. The contrast and brightness of the images are enhanced for visibility.}
    \label{fig:TIP_example}
\end{figure}

\begin{enumerate}
    \item \textbf{Tail intersection point (TIP)}: This representation is calculated in three steps using three fins and the bounding box center of the fish body:
\begin{enumerate}
    \item Find the center of the anal fin (ANF), adipose fin (APF), tail fin (TF) and body (B).
    \item Find the intersection point (IP) between the two line segments $[ANF, APF]$ and $[TF, B]$.
    \item Calculate the length ratio of the two line segments $[APF, IP]$ and $[APF, ANF]$. This provides a number, the TIP, in the range $[0, 1]$ that describes the salmon tail beat state (\cref{fig:TIP_example}).
\end{enumerate}
\item \textbf{Tailfin width}: The width of the tail fin bounding box.
\item \textbf{Salmon width}: The width of the salmon bounding box.
\end{enumerate}

Extracting a salmon tail beat state for a sequence of frames provides a time series of salmon tail beat states. This time series was smoothed by a Savitzky-Golay filter, with a window length of 11 frames and a polynomial order of 2. The maxima and minima of this smoothed signal were found as local extrema by the SciPy \cite{2020SciPy} $find\_peaks()$ method. The distance between two maxima or two minima is interpreted as the current tail beat wavelength of the salmon.

\section{Experiments}
\label{sec:results}
\subsection{Crowded salmon tracking}
%\subsubsection{Evaluation metrics}
We evaluated the four models trained on the CS train dataset (M1-M4) on the CS validation dataset. Three types of events were measured/counted using the TrackEval \cite{Luiten2020} software: I) \textbf{ID Transfers} (when the same tracker ID follows different fish over time; e.g., Tracker 1 follows Fish A in one frame and Fish B in another), II) \textbf{ID Switches} (when the same fish is assigned different tracker IDs over time; e.g., Fish A is matched to Tracker 1 in one frame and to Tracker 2 in the next) and III) \textbf{Matches} (total number of matches). In the ideal case, the tracker would have no ID transfers or ID switches, and the same number of matches as the ground truth dataset (871 salmon and 5814 body parts, see CS val in \cref{tab:datasets}). Since we considered each body part separately in our trackers, we modified the TrackEval software to only consider hypothesis-object matches of the same class. 

%\subsubsection{Experimental setup}
The original BT method includes a \textit{camera motion compensation} (CMC) module, but we observed similar performance and simplified processing line by not activating this module. 
Furthermore, we disabled the \textit{appearance embedding} to avoid data gathering biases when employing the tracker for the construction of salmon re-identification datasets. The $min\_hits$ parameter (see the BT documentation \cite{Stanojevic2024}) was set to zero.

\begin{figure}
    \centering
    \includegraphics[width=\linewidth]{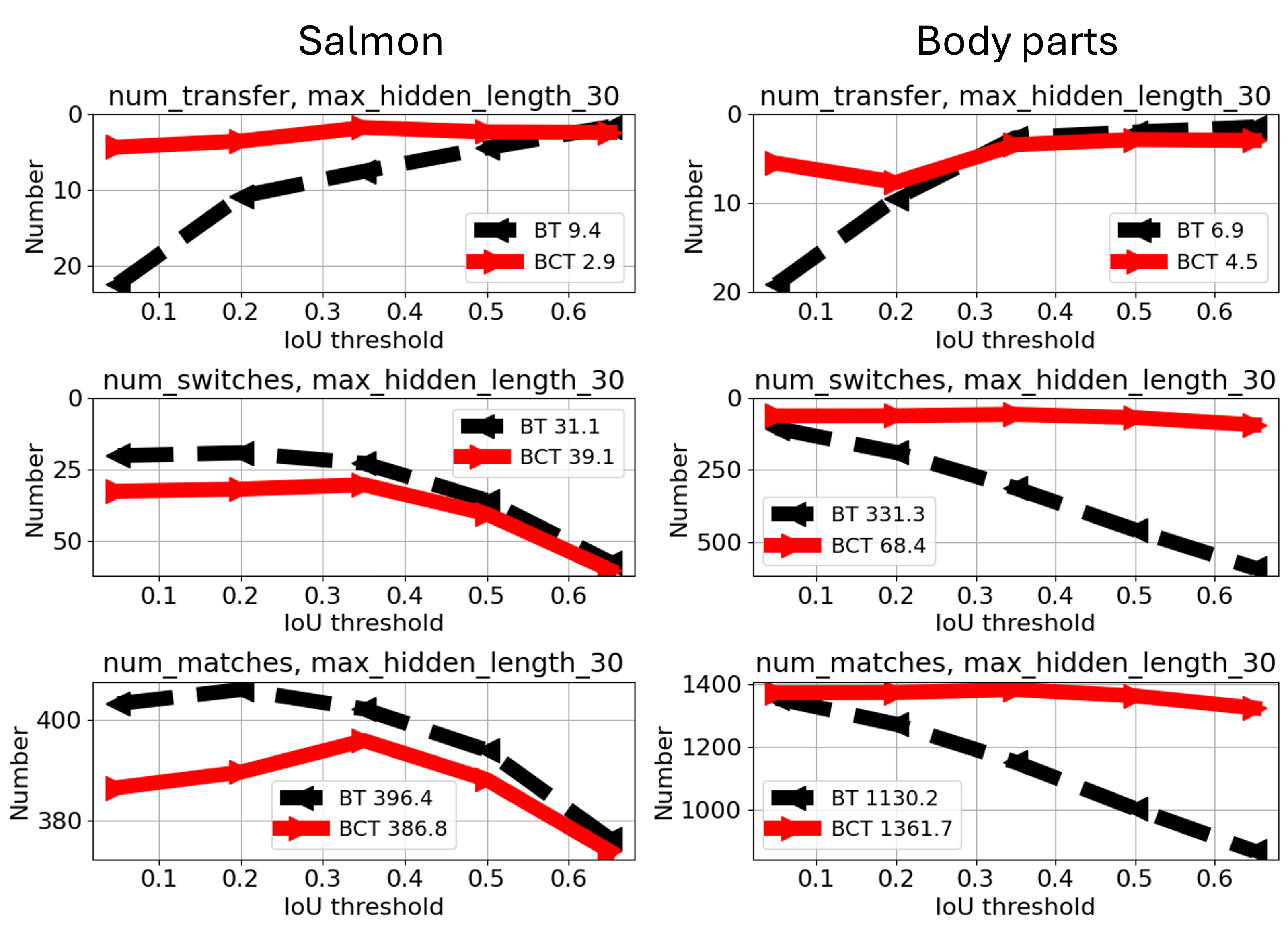}
    \caption{Number of salmon events and body part events for different IoU thresholds. Note that the y-axis range of the plots changes since each plot displays a different event and object group (salmon and body parts). \textit{BT} = BoostTrack, \textit{BCT} = BoostCompTrack.}
    \label{fig:CS_res}
\end{figure}

We compared both the salmon events and the salmon body part events for different association types (BT and BCT), detectors (M1-M4), IoU thresholds (0.05, 0.2, 0.35, 0.50 and 0.65), and \textit{hidden lengths} (3 and 30). 
%%% NEW 2.9 %%%
By IoU threshold, we mean the minimum overlap (as modified in \cite{Stanojevic2024}) required to match a tracker prediction to a detection. Thresholds above 0.65 offered minor transfer reductions while significantly increasing switch rates, rendering them impractical (\cref{fig:CS_res}).
%%%
\textit{Hidden length} is the maximum number of frames a tracker is allowed to go unmatched before it is deleted. Longer \textit{hidden lengths} help maintain tracks through occlusions or missed detections (reducing ID switches), but can increase ID transfers by linking different objects over time. 

%\subsubsection{Results and Observations}
%The evaluation events (ID transfers, ID switches and matches) averaged over the evaluated thresholds are presented in \cref{tab:results_CS}, while \cref{fig:CS_res} displays how the salmon and salmon body part events change as a function of IoU threshold for fixed model M3 and a fixed \textit{hidden length} setting of 30.

\cref{tab:results_CS} presents evaluation events (ID transfers, ID switches, and matches) averaged over all thresholds, while \cref{fig:CS_res} shows how salmon and body-part events vary with IoU threshold for model M3 and a fixed \textit{hidden length} of 30.

\begin{table}[h]
\centering
\small

\captionsetup[subtable]{font=small,skip=2pt,justification=centering,singlelinecheck=false}

\begin{subtable}[t]{\linewidth}
\caption{Hidden length 3}
\label{tab:CS_results_hl3}
\begin{tabular}{|p{2.26cm}|c|c|c|c|c|c|}
\hline
 & \multicolumn{2}{c|}{Transfer $\downarrow$} & \multicolumn{2}{c|}{Switches $\downarrow$} & \multicolumn{2}{c|}{Matches $\uparrow$} \\
 & \textit{sal} & \textit{bp} & \textit{sal} & \textit{bp} & \textit{sal} & \textit{bp} \\
\hline
BT M1 & 6.9 & 7.1 & 43 & 270 & 341 & 823 \\ \hline
BCT all M1 & 1.5 & 6.4 & 56 & \cellcolor{green!10}\textbf{89} & 325 & 999 \\ \hline \hline
BT M3 & 5.5 & 2.5 & 42 & 348 & 387 & 1115 \\ \hline
BCT all M3 & \cellcolor{green!10}\textbf{1.4} & 5.1 & 45 & 90 & 381 & 1356 \\ \hline
\hline
BT M2 & 7.2 & 8.3 & 37 & 150 & 337 & 721 \\ \hline
BT M4 & 3.4 & 2.7 & 28 & 156 & 370 & 860 \\ \hline
\end{tabular}
\end{subtable}

\vspace{0.5em}

\begin{subtable}[t]{\linewidth}
\caption{Hidden length 30}
\label{tab:CS_results_hl30}
\begin{tabular}{|p{2.26cm}|c|c|c|c|c|c|}
\hline
 & \multicolumn{2}{c|}{Transfer $\downarrow$} & \multicolumn{2}{c|}{Switches $\downarrow$} & \multicolumn{2}{c|}{Matches $\uparrow$} \\
 & \textit{sal} & \textit{bp} & \textit{sal} & \textit{bp} & \textit{sal} & \textit{bp} \\
\hline
BT M1 & 9.2 & 11.2 & 36 & 260 & 348 & 842 \\ \hline
BCT all M1 & \cellcolor{green!10}\textbf{2.1} & 7.1 & 52 & 76 & 327 & 999 \\ \hline \hline
BT M3 & 9.4 & 6.9 & 31 & 331 & 396 & 1130 \\ \hline
BCT all M3 & 2.9 & 4.5 & 39 & \cellcolor{green!10}\textbf{68} & 387 & 1362 \\ \hline
\hline
BT M2 & 8.9 & 12.2 & 35 & 137 & 339 & 734 \\ \hline
BT M4 & 5.8 & 7.6 & 20 & 135 & 378 & 880 \\ \hline
\end{tabular}
\end{subtable}

\vspace{0.5em}

\begin{subtable}[t]{\linewidth}
\caption{Ablation}
\label{tab:CS_results_ablation}
\begin{tabular}{|p{2.26cm}|c|c|c|c|c|c|}
\hline
 & \multicolumn{2}{c|}{Transfer $\downarrow$} & \multicolumn{2}{c|}{Switches $\downarrow$} & \multicolumn{2}{c|}{Matches $\uparrow$} \\
 & \textit{sal} & \textit{bp} & \textit{sal} & \textit{bp} & \textit{sal} & \textit{bp} \\
\hline
BT M3 & 5.5 & 2.5 & 42 & 348 & 387 & 1115 \\ \hline
BT CMC M3 & 5.6 & 2.5 & 42 & 339 & 387 & 1123 \\ \hline
BCT M3 & 5.5 & 14.8 & 42 & 97 & 387 & 1362 \\ \hline
BCT turn M3 & 5.5 & 14.8 & 42 & 97 & 387 & 1362 \\ \hline
BCT bpiou M3 & 5.5 & 14.4 & 42 & 93 & 387 & 1356 \\ \hline
BCT bpdis M3 & 3.9 & 10.9 & 41 & 92 & 387 & 1363 \\ \hline
BCT nobp M3 & 2.5 & 7.9 & 47 & 100 & 381 & 1357 \\ \hline
BCT all CMC M3 & 1.8 & 6.7 & 44 & 91 & 383 & 1352 \\ \hline
BCT all M3 & \cellcolor{green!10}\textbf{1.4} & 5.1 & 45 & \cellcolor{green!10}\textbf{90} & 381 & 1356 \\ \hline
\end{tabular}
\end{subtable}

\caption{Evaluation events 
%%% NEW 1.7, 2.1 %%%
on the CrowdedSalmon (CS) validation dataset 
%%%
for salmon (left column of each event) and salmon body parts (right column of each event) averaged over thresholds between 0.05 and 0.65. The lowest salmon ID transfer number and the lowest salmon body part ID switch number for each table are marked in green. BCT is an abbreviation for \textit{BoostCompTrack} (our novel method), BT for BoostTrack and CMC for \textit{camera motion compensation}.}
\label{tab:results_CS}
\end{table}

The pose models (M1, M3) yielded more matches than the bounding box models (M2, M4) due to better detection of small parts such as the adipose fin. However, they also showed an increased number of ID switches because of poorer object localization.
Since the pose models employ keypoint losses and keypoint pretraining, it is not surprising that using the pose coordinates as box corners leads to subpar localization accuracy. 

%%% NEW 2.11 %%%
%These localization errors were significantly reduced in M5 (qualitatively observed when evaluated on different video sequences), suggesting that the limited size of the CS train set significantly limits the performance of M1 and M3. 
%%%

% Complete model
For the final BCT model (BCT all), we observe a significant reduction in salmon ID transfers as the \textit{CrowdedModule} (including bpdis, nobp and bpiou) uses the knowledge about the body parts to discard poor associations. Furthermore, the body part ID switches are reduced significantly due to body parts retaining their ID during occlusion as long as part of the salmon is visible. Both of these observations are highlighted in \cref{tab:results_CS}, by coloring the lowest salmon ID transfer counts and body part ID switch counts in green. The number of salmon matches is slightly reduced for BCT since our check modules cause some salmon associations to be discarded. The body part matches, however, increase due to situations where previously unassociated body parts are correctly matched because of salmon associations.

% medium/nano and hidden length
The discrepancy in salmon ID switches and matches between BT and BCT is larger for the nano pose model than the medium one, which is due to less accurate keyboxes leading to more rejected associations by the \textit{CrowdedModule}. Furthermore, we observe that using a longer hidden length (30) increases ID transfers and reduces ID switches for both salmon and body parts, as expected from the greater uncertainty introduced by longer association gaps.

%\subsubsection{Ablation study}
%%% NEW shortened %%%
An ablation of the BCT modules (\textit{TurnModule}, bpdis, nobp and bpiou) using M3 with \textit{hidden length} 3 (\cref{tab:CS_results_ablation}) showed that bpdis and nobp effectively reduced salmon ID transfers, especially at low IoU thresholds (\cref{fig:CS_res} is representative). Of the two, nobp triggered more often, leading to greater match loss and more ID switches. The bpiou module caused a reduction in both body part ID transfers and body part ID switches by removing body part detections of poor quality. Switching from individual-based (BT) to salmon-based (BCT) associations initially caused an increase in body part ID transfers due to propagated salmon errors, but enabling all modules balanced the body part ID transfer performance.
%%%

%%% NEW 1.8, 1.9 %%%
Analysis of the generated tracks indicates that BCT tracking errors primarily result from inaccurate keypoint localization, distant fish, and occlusions. While full occlusions remain challenging, our system performs reliably under partial occlusion. For nearby fish that are well-represented in the training distribution, the system achieves sufficient accuracy for most practical applications.

\begin{figure}
    \centering
    \includegraphics[width = \linewidth]{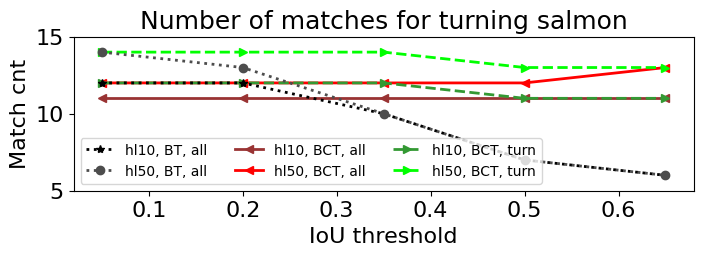}
    \caption{Number of matches for turning salmon in the TurnSalmon validation dataset. \textit{hl} indicates the \textit{hidden length} setting. \textit{BT} = BoostTrack, \textit{BCT} = BoostCompTrack. \textit{all} uses all BCT modules; \textit{turn} uses only the TurnModule. 
    %%% NEW 2.10 %%%
    The horizontal axis specifies the IoU setting of the tracker.
    %%%%
    }    \label{fig:turnsalmon_matches}
\end{figure}

\subsection{Turning salmon tracking}
We compared how well our method was able to maintain the track of turning salmon. The events of the keybox detection model (M5, see \cref{sec:method_salmon_detection}) evaluated on the TS dataset were gathered by comparing the tracking (hypothesis) ID associated with the two labels in each ground truth track, where the two labels refer to the two sides of the fish (\cref{sec:dataset}).
%%% NEW 2.1, 2.10 %%%
Each ground truth bounding box was matched to the tracker hypothesis bounding box with the highest overlap, requiring a minimum IoU of 0.2.
%%%

In \cref{fig:turnsalmon_matches}, we see the number of matches of the turning ground truth tracks for BT and BCT under different thresholds, \textit{hidden lengths} and included modules (all or only the \textit{TurnModule}). We observe that all BCT trackers approximately maintain their initial match count for all IoU thresholds ($^+_- 1$ match). This suggests that the tracker precisely detects which salmon are turning, and reduces the association requirements for these tracks. The \textit{TurnModule} did not impact the CS validation results (\cref{tab:CS_results_ablation}), suggesting the module avoids impacting tracks without turning salmon. We observe that the full BCT tracker has slightly lower match counts; this is due to the \textit{CrowdedModule} increasing the ID switch count (by terminating poor trackers), which causes a reduction in match count. 

\subsection{Tail beat wavelength monitoring}
To evaluate the quality of the tail beat wavelength estimation, we compared the estimated extrema of the time series with the ground truth tail beat state extrema. Since the wavelength varied significantly, we estimated the proximity of tail beat state extrema rather than a single frequency number. To match ground truth and hypothesis (time series) extrema, we employed a greedy strategy where we picked the closest matches (hypothesis and ground truth with similar frame numbers) first, until all matches below a threshold were assigned. We calculated events (true positive, false positive and false negative) for all proposed tail beat state representations, averaged over thresholds between one and four frames. To ensure a fair comparison, we tuned the prominence parameter of the $find\_peaks()$ function to yield approximately the same number of extrema as the number of ground truth extrema. This tuning was performed by a binary search (running for 50 iterations) on the prominence parameter.
%%% NEW %%%
% One sample has 3 in difference. All other have 1 or 0. Therefore, removing this sentence.
%resulting in an extrema count difference of at most 1.
%%%

\begin{table}[h!]
\centering
\begin{tabular}{|l|c|c|c|}
\hline
 & TP $\uparrow$ & FP $\downarrow$ & FN $\downarrow$ \\
\hline
Salmon width    & 42.50 & 56.50 & 56.50 \\
Tailfin width   & 63.25 & 37.75 & 35.75 \\
TIP             & \cellcolor{green!10}\textbf{66.00} & \cellcolor{green!10}\textbf{34.00} & \cellcolor{green!10}\textbf{33.00} \\
\hline
\end{tabular}
\caption{True positives (TP), false positives (FP) and false negatives (FN) for different tail beat state representations, averaged over frame thresholds between one and four. 
%%% NEW 1.3 %%%
A TP was defined as a matched pair of estimated and ground truth extrema within the threshold; an FP was an unmatched estimated extremum; and an FN was a ground truth extremum with no corresponding match. 
%%%
The best result for each event type is highlighted in green.}
\label{tab:results_tailbeat}
\end{table}

\begin{figure}
    \centering
    \includegraphics[width=\linewidth]{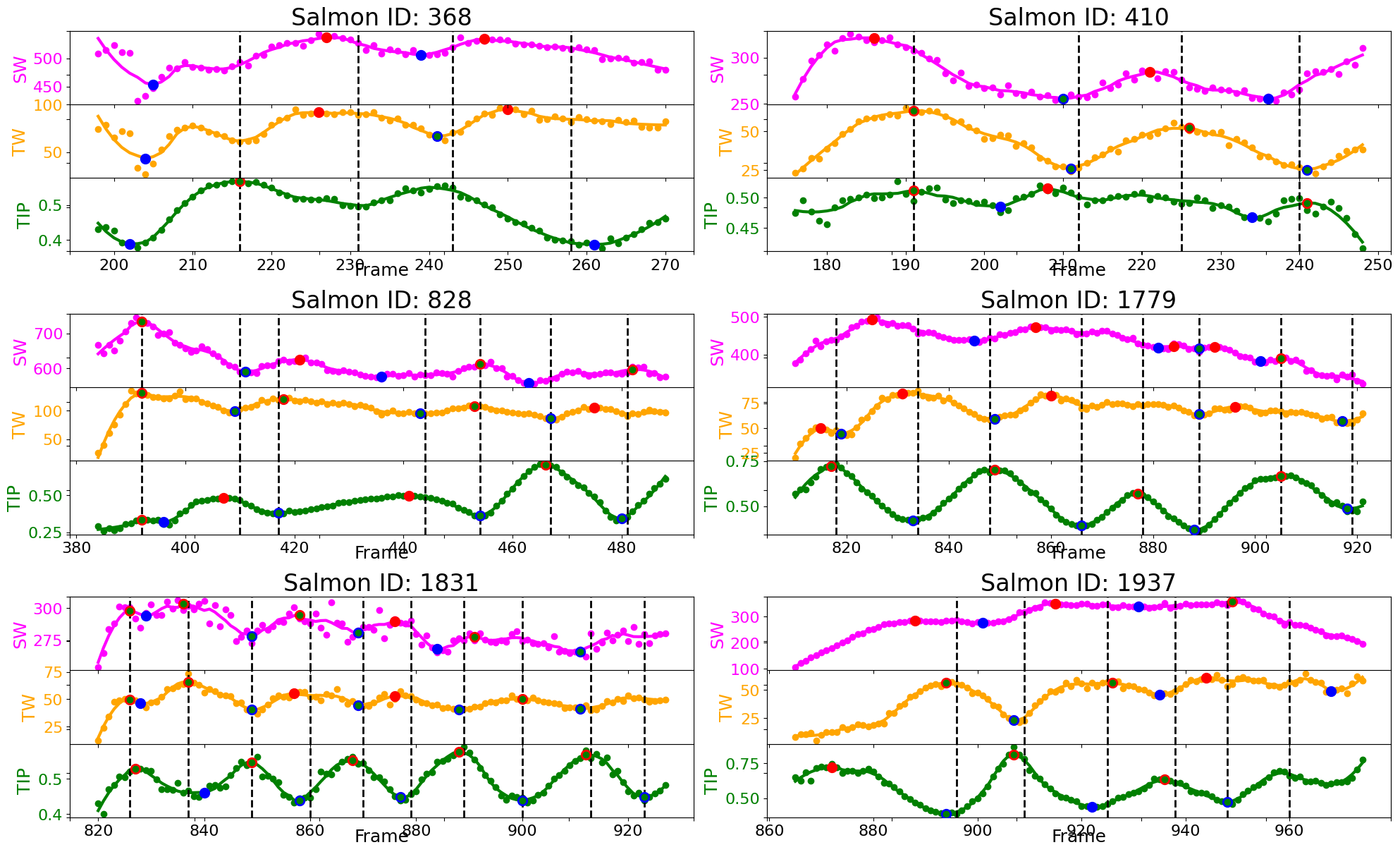}
    \caption{Visualized results from the salmon tail beat wavelength experiment. Tail beat state values are shown as time series for three representations: \textit{Salmon width} (SW), \textit{Tailfin width} (TW), and \textit{Tail intersection point} (TIP). Ground-truth extrema appear as black vertical lines, with smoothed peaks in red, smoothed lows in blue, and correctly classified extrema (within two frames) marked by a small green dot inside the corresponding extrema.%Visualized results from the salmon tail beat wavelength experiment. Tail beat state values are represented as a time series for three different representations: \textit{Salmon width} (SW), \textit{Tailfin width} (TW) and \textit{tail intersection point} (TIP). Ground truth extrema are represented as black, vertical lines, smoothed time series peaks as red dots, smoothed time series lows as blue dots, and tail beat state extrema classified as correct, given a frame threshold of two, marked by a small green dot inside their larger extrema dots.
    }
    \label{fig:tail_beat_ts}
\end{figure}

%The results of the different tail beat state representations are shown in \cref{tab:results_tailbeat}, while the tail beat state time series are displayed in \cref{fig:tail_beat_ts}. It is clear from both the quantitative results and by looking at the time series that the TIP and \textit{Tailfin width} representations are superior to the  \textit{Salmon width} representation. 

%%% New shortened %%%
As shown in \cref{tab:results_tailbeat} and \cref{fig:tail_beat_ts}, both the TIP and \textit{Tailfin width} representations yield superior performance compared to \textit{Salmon width}. 
%%%
This demonstrates that our tracker, which facilitates all representations, is better suited for salmon tail beat frequency estimation than trackers relying on salmon bounding boxes (which can only use the  \textit{Salmon width} representation). Furthermore, our tracker offers more flexibility than pose networks (which can only use the  \textit{Salmon width} or TIP representations).

%The TIP and \textit{Tailfin width} representations achieve approximately the same results.
%One weakness of the TIP representation is that a salmon above the camera will have a camera-facing tail arching projected high in the frame, whereas a salmon at the bottom of the frame will have a camera-facing tail arching projected low in the frame. This leads to a 180-degree phase shift in the tail beat state time series if a salmon swims from above the camera to below the camera, 
%%% NEW %%%
%and issues when the fish are swimming at this line (ID 410 in \cref{fig:tail_beat_ts}).
%%%
%This is not the case for the \textit{Tailfin width} representation. On the other hand, with the proper logic implemented, the TIP representation can distinguish tail beat motion towards and away from the camera, while the \textit{Tailfin width} representation cannot do this.

%%% New shortened %%%
The TIP and \textit{Tailfin width} representations yield comparable results. A limitation of TIP is its sensitivity to vertical fish position: when a salmon moves from above to below the camera, the tail arching appears inverted, causing a 180-degree phase shift in the time series. Near this transition, the representation may break down (e.g., ID 410 in \cref{fig:tail_beat_ts}). This issue does not affect the \textit{Tailfin width} representation. However, TIP can, with appropriate logic, distinguish tail motion direction relative to the camera, something \textit{Tailfin width} is less suited to handle reliably.
%%%

%%% New paragraph, 1.4 %%%
The primary sources of error for the best representation (TIP) include minor misalignments between detected peaks and ground truth extrema, as well as false positives at the time series boundaries, which can lead to false negatives elsewhere due to matched peak numbers (e.g., ID 1937 in \cref{fig:tail_beat_ts}). Improved peak detection algorithms without peak count matching could mitigate these false negatives. Given the inherent difficulty of manually annotating tail beat extrema, some discrepancies may stem from ground truth uncertainty. The automated approach may rival the precision of these manual labels.

%Ground truth uncertainty also plays a role, as manual annotation of tail beat extrema is inherently challenging. Our automated approach may match the accuracy of the ground truth annotations.

\section{Conclusion}
This paper introduced a flexible framework for salmon tracking in support of welfare monitoring in complex underwater environments. We evaluated its performance on two novel datasets, demonstrating that our method outperforms BoostTrack in crowded underwater scenes and in tracking turning salmon. Furthermore, we showed that the detailed body part tracks produced by our approach can be directly used for tail beat wavelength estimation, a key welfare indicator.

To emphasize the adaptability of our framework, we deliberately relied on small datasets and lightweight modules. Despite using fewer than 100 labeled frames, the system achieved accurate tracking of both whole salmon and individual body parts across diverse and challenging scenarios. These results highlight the versatility of our method and its potential for broader application with minimal adjustments.

Looking ahead, we aim to extract additional welfare indicators from the informative tracks generated by our system, moving toward a comprehensive, low-overhead monitoring solution. By linking body parts, our method supports the integration of multiple indicators at the level of individual fish, opening the door to long-term health monitoring, enabled by consistent re-identification through analysis of distinct texture regions such as fins, body, and head.

\section*{Acknowledgements}
This work was supported by the Research Council of Norway (RCN) under project number 344022, titled ’\textit{Computer Vision and Artificial Intelligence-based Salmon Identification and automated long-term welfare assessment in aquaculture net-pens (cAIge)}'.
{
    \small
    \bibliographystyle{ieeenat_fullname}
    \bibliography{main}
}

\end{document}